\pdfoutput=1
\documentclass[11pt]{article}
\usepackage[preprint]{acl}

\usepackage{times}
\usepackage{latexsym}
\usepackage{booktabs}  
\usepackage{algorithm}
\usepackage{algpseudocode}  
\usepackage{amsmath} 
\usepackage{amssymb}
\usepackage{tabularx}
\usepackage{hyperref}
\algrenewcommand\algorithmicrequire{\textbf{Input:}}
\algrenewcommand\algorithmicensure{\textbf{Output:}}
\usepackage[T1]{fontenc}
\usepackage[utf8]{inputenc}
\usepackage{microtype}
\usepackage{inconsolata}
\usepackage{graphicx}
\title{Can LLMs Simulate Social Media Engagement? A Study on Action-Guided Response Generation}

\author{
 \textbf{Zhongyi Qiu\textsuperscript{1}},
 \textbf{Hanjia Lyu\textsuperscript{2}},
 \textbf{Wei Xiong\textsuperscript{2}},
 \textbf{Jiebo Luo\textsuperscript{2}}
\\
 \textsuperscript{1}School of Computational Science and Engineering, Georgia Institute of Technology \\
 \textsuperscript{2}Department of Computer Science, University of Rochester
\\
 \texttt{zhongyiqiu@gatech.edu, hlyu5@ur.rochester.edu} \\
 \texttt{wxiongur@gmail.com, jluo@cs.rochester.edu}
}

\begin{document}
\maketitle
\begin{abstract}
Social media enables dynamic user engagement with trending topics, and recent research has explored the potential of large language models (LLMs) for response generation. While some studies investigate LLMs as agents for simulating user behavior on social media, their focus remains on practical viability and scalability rather than a deeper understanding of how well LLM aligns with human behavior. 
This paper analyzes LLMs' ability to simulate social media engagement through action-guided response generation, where a model first predicts a user's most likely engagement action—retweet, quote, or rewrite—towards a trending post before generating a personalized response conditioned on the predicted action. 
We benchmark {\tt GPT-4o-mini}, {\tt O1-mini}, and {\tt DeepSeek-R1} in social media engagement simulation regarding a major societal event discussed on $\mathbb{X}$.
Our findings reveal that zero-shot LLMs underperform BERT in action prediction, while few-shot prompting initially degrades the prediction accuracy of LLMs with limited examples. However, in response generation, few-shot LLMs achieve stronger semantic alignment with ground truth posts.
\end{abstract}

\section{Introduction}
Social media has transformed how users engage with information, enabling rapid discussions on trending topics through various forms of interaction~\cite{miller2016world, roth2022viral}. Research on how people engage with trending posts has become a prominent topic, as these interactions influence public discourse and offer valuable insights into online behavioral patterns~\cite{van2013understanding}. Meanwhile, large language models (LLMs) have been widely used for understanding social media posts, generating content, and simulating user behavior~\cite{y2022large,jiang2023social,lyu2024gpt}.  

\begin{figure}[t]
\centering
\includegraphics[width=\linewidth]{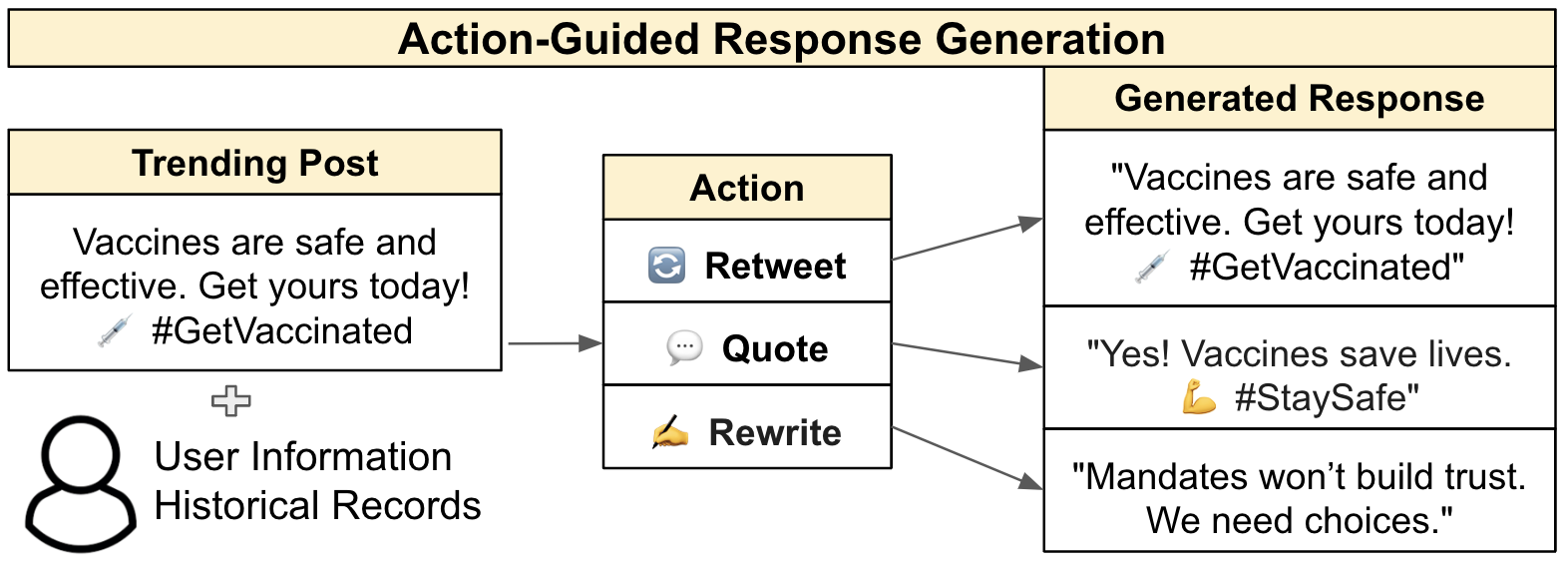}
\caption{An illustration of action-guided response generation: The model first predicts the engagement action in response to the trending post. The post is then generated based on the predicted action.}
\vspace{-3mm}
\label{fig:idea_overview}
\end{figure}

Recent research has increasingly leveraged LLMs to generate human responses to trending posts~\cite{yu2024popalm, salemi2025reasoning}. To better simulate real social media interactions, researchers have begun incorporating user behaviors---such as retweeting, quoting, or posting new comments~\cite{yu2024repalm}---into the post-generation process. However, existing studies primarily focus on the feasibility and scalability of LLMs as agent-based simulators~\cite{mou2024unveiling, cai2024language,gao2023s}, without conducting an in-depth analysis of their ability to capture user engagement dynamics or establishing robust evaluation frameworks. 

To analyze the capability of LLMs as simulators in predicting user engagement with trending posts and generating corresponding responses, we follow an action-guided response generation framework, as illustrated in Figure~\ref{fig:idea_overview}. In this framework, we categorize user actions into three major types: retweet, quote, and rewrite~\cite{pillai2025engagement}. We first predict a user’s engagement type and then use this prediction to guide the response generation process. This structured process captures user engagement dynamics, enabling more informed response generation~\cite{toraman2022understanding}. 

We evaluate the simulation quality by measuring the alignment between the generated posts and the real user posts on two high-level aspects: \textbf{semantics} and \textbf{style}.
Specifically, we assess semantics-preserving consistency, which examines whether the generated response preserves key attributes such as sentiment, emotion, stance, age, and gender~\cite{10825696}.
For style alignment, we analyze the linguistic attributes using LIWC\footnote{https://www.liwc.app} and compute the cosine similarity to quantify textual resemblance. 

We benchmark the performance of a few representative LLMs, including {\tt GPT-4o-mini}, {\tt O1-mini}, and {\tt DeepSeek-R1}. Furthermore, we analyze the impact of incorporating user information and historical behavior for user engagement simulation.

Our findings show that zero-shot LLMs underperform BERT in action prediction. Moreover, while few-shot prompting initially degrades action prediction accuracy, performance improves as the number of shots increases. Despite this, few-shot LLMs generate responses better aligned with the semantics of ground-truth posts.

\section{Related Work}
AI-driven social media content generation has been explored across various domains. \citet{lim2023artificial} studied prompt engineering for health awareness messages, showing AI-generated content can match or exceed human quality, but their work focused on general messages rather than personalized engagement. \citet{yu2024repalm} introduced RePALM, a model optimizing quote tweets for likes, yet it did not account for diverse engagement types like retweets and rewrites or user-specific behaviors. \citet{rossetti2024social} introduced Y SOCIAL, a digital twin of a physical social media platform, but it lacks a direct comparison with real-world social media interactions. \citet{dmonte2024classifying} proposed HiSim, a hybrid scalable framework for social media simulation, but it lacks a detailed step-by-step accuracy analysis of LLM-driven user simulation. 

While these studies primarily focus on the feasibility and scalability of LLM-driven social media simulations, our work takes a different approach by focusing on examining the performance of LLMs in capturing user engagement dynamics.
Rather than optimizing for engagement metrics or developing large-scale simulation frameworks, we aim to understand how well LLMs align with human behavior in social media interactions, investigating both their successes and limitations in action-guided response generation.

\section{Experiments}

\subsection{Data Precessing}
We benchmark the simulation performance on a dataset of tweets discussing COVID-19 vaccination on $\mathbb{X}$ (formerly Twitter) from September 28, 2020, to November 4, 2020~\cite{lyu2022social}. This initial dataset contains 6.3 million entries. After removing special characters, URLs, and other non-informative tokens, we locate the original tweet for every retweet and quote in the dataset. We then calculate the total engagement for each original tweet by summing its retweet and quote counts. Next, to focus on the most influential vaccination-related content, we filter the dataset to obtain the top 200 tweets with the highest total engagement (i.e., retweet + quote counts). We denote these as our “trending tweets”.

After identifying the trending posts, we examine three types of user engagement: (1) retweets, (2) quote tweets, and (3) rewrites. The first two categories are readily extracted from the dataset via their respective labels. For rewrites---where a user composes an entirely new tweet discussing the same topic without directly quoting or retweeting the popular tweet---we design a dedicated pipeline. Specifically, for each popular tweet, we filter all tweets posted within the subsequent seven days, convert their text into embeddings, and then apply cosine similarity to retain only those tweets closely matching the content of the original popular tweet. Finally, we perform keyword matching to ensure topic alignment, producing a set of “rewrite” tweets that address the same vaccination-related context without reference to the original.

After extracting the rewrite data, we balanced the dataset. Our final dataset has 3,990 entries, consisting of 1,330 quote posts, 1,330 retweets, and 1,330 rewrites.

\begin{table*}[t]
\centering
\small %
\vspace{-2mm}
\caption{Accuracy of action prediction and alignment of generated responses with ground truth in terms of semantics and style across different models and configurations.}
\label{tab:action_text_performance}
\vspace{-1mm}
\begin{tabular}{llc|cccccc}
\toprule
\multicolumn{3}{c|}{\textbf{Action Prediction}} & \multicolumn{6}{c}{\textbf{Text Generation}} \\
\midrule
\textbf{Model} & \textbf{Config.} & \textbf{Act. Acc.}  & \textbf{Sent.} & \textbf{Emo.} & \textbf{Stance} & \textbf{Age} & \textbf{Gender} & \textbf{LIWC} \\
\midrule
GPT-4o-mini  & ZS\_base  & 33.78\%  & 49.70\% & 44.69\% & 48.82\%  & \textbf{98.42\%}  & 48.07\%  & 0.7978 \\
             & ZS\_ info & \textbf{42.23\% } & 50.28\% & 45.61\% & 50.05\%  & 95.96\%  & 53.71\% & \textbf{0.8151} \\
             & FS\_ hist & 28.75\%  & \textbf{62.26\%} & \textbf{57.89\%} & \textbf{63.28\%}  & 96.59\%  & \textbf{58.87\%} & 0.8105 \\
\midrule
O1-mini      & ZS\_ base  & 33.15\%  & 47.39\% & 38.97\% & 45.51\%  & 82.76\%  & 56.52\% & 0.7433 \\
             & ZS\_ info  & \textbf{52.76\%} & 55.74\% & 47.74\% & 56.37\%  & 80.30\%  & 68.70\%  & 0.7877 \\
             & FS\_ hist  & 44.21\%  & \textbf{63.51\%} & \textbf{57.49\%} & \textbf{65.01\%}  & \textbf{84.64\% } & \textbf{70.13\% } & \textbf{0.7950} \\
\midrule
DeepSeek-R1  & ZS\_ base  & 35.41\%  & 63.13\% & 59.30\% & 62.41\%  & 97.14\%  & 58.02\%  & 0.7685 \\
             & ZS\_ info  & \textbf{53.18\%} & 62.61\% & 57.64\% & 63.23\%  & 82.01\%  & 69.72\%  & \textbf{0.7832} \\
             & FS\_ hist & 33.66\%  & \textbf{68.85\%} & \textbf{62.88\%} & \textbf{70.05\% } & \textbf{87.94\%}  & \textbf{70.20\%} &  0.7286\\
\midrule
Bert+GPT-4o-mini & Frozen & 46.09\% & \textbf{69.02\%} & \textbf{60.03\%} & \textbf{69.40\%} & \textbf{94.21\%} & \textbf{66.44\%} & \textbf{0.8143} \\
             & Fine-Tuned & \textbf{57.14\%} & 55.66\% & 46.14\% & 56.27\%  & 93.96\%  & 57.63\%  & 0.7798 \\

\bottomrule  

\end{tabular}%

\vspace{2mm}

\small ZS\_base = Zero-shot Baseline. 
ZS\_info = Zero-shot with User Information. 
FS\_hist = Few-shot with User Information and Historical Examples. 
Act. Acc. = Action Accuracy. 
Text Generation values = accuracy against ground truth for each attribute.
LIWC values = cosine similarity between the LIWC feature vectors of generated and ground-truth posts.
\vspace{-2mm}

\end{table*}

\begin{figure}[t]
    \centering
    \includegraphics[width=0.45\textwidth]{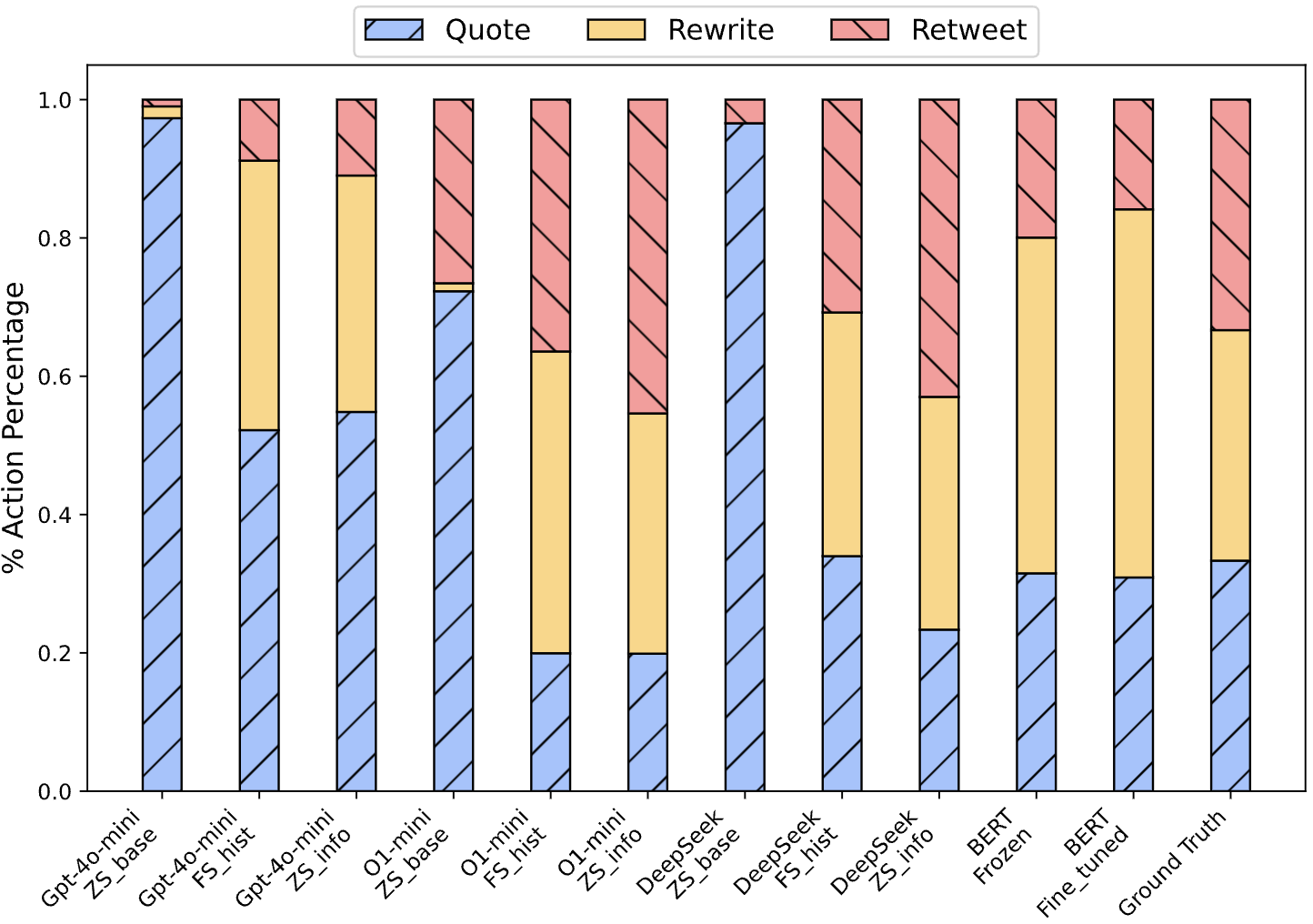}
    \caption{Statistics of quote, rewrite, and retweet actions predicted by different models. 
    The models are grouped into Gpt-4o-mini, O1-mini, DeepSeek, and BERT, with three settings: ZS\_base, ZS\_info, and FS\_hist. 
    Ground truth is also included as a reference.}
    \label{fig:action_distributions}
    \vspace{-4mm}
\end{figure}

\subsection{Experimental Setup}

Each data instance consists of a user and a trending post. The goal is to predict the user's engagement action—retweet, quote, or rewrite. If the predicted action is quote or rewrite, a generative model is invoked to generate the user’s response. Otherwise, the post remains unchanged. Details of the specific algorithms are provided in Appendix~\ref{sec:algorithm}.

We conduct experiments under three configurations for {\tt GPT-4o-mini}, {\tt O1-mini} and {\tt DeepSeek-R1-671b}\footnote{For simplicity, we refer to it as {\tt DeepSeek-R1}.}: (1) Zero-shot Baseline (ZS\_base), where LLMs are given only the content of the trending tweet; (2) Zero-shot with User Information (ZS\_info), in which LLMs also receive basic user profile data; and (3) Few-shot with User Information and Historical Examples (FS\_hist), where LLMs have access to both user information and examples of the user’s past reactions to similar popular tweets. 

For comparison, we evaluate two BERT-based models for action prediction: (1) a frozen BERT model with only the classifier layer trained and (2) a fully fine-tuned BERT model. Both models are trained as three-way classifiers (retweet, quote, rewrite) using a labeled dataset that incorporates user-specific features and historical behavioral patterns. Once the model predicts the user’s action, we use {\tt GPT-4o-mini} to generate tweet content for quote or rewrite cases.

\subsection{Evaluation Setting}
We assess our methods along two dimensions: action prediction and post generation. For action prediction, we compute the overall accuracy of identifying whether the user will retweet, quote, or rewrite a given popular tweet. In the tweet generation stage, we evaluate how closely the generated tweets align with ground-truth tweets in terms of both semantics preservation and stylistic consistency. The attributes used to measure semantics and style are detailed in Appendix~\ref{sec:attributes}.  

We use {\tt Llama-3.3-70B} to label the semantics and style of each tweet (both generated and ground truth). Inspired by \citet{lyu2024computational}, we use Linguistic Inquiry and Word Count (LIWC) to derive vector representations of both generated and ground-truth tweets. By computing similarities between these vectors, we quantify style alignment.

\section{Results}

Table~\ref{tab:action_text_performance} summarizes the performance of different LLMs and configurations. Through the analysis on these results, we reveal three key observations.

\subsection{Fine-tuned BERT outperforms Zero-shot LLMs in action prediction}
Our results show that the fine-tuned BERT model achieves the highest performance for action prediction compared to Zero-shot LLM-based methods. Specifically, Fine-tuned BERT achieves an accuracy of 57.14\%, surpassing the best-performing Zero-shot LLM-based method, {\tt DeepSeek-R1} (ZS\_info), which achieves 53.18\%. 

To understand why LLMs underperform in action prediction, we examine their prediction distributions. Figure~\ref{fig:action_distributions} shows the statistics of actions predicted by {\tt GPT-4o-mini}, {\tt O1-mini} and {\tt Deepseek-R1}, in three settings: ZS\_base, ZS\_info and FS\_hist as well as Frozen Bert and Fine-tuned Bert. We observe that \textbf{LLMs predominantly predicts the quote action, indicating a strong bias toward this category}. After incorporating user information and historical examples, this bias is reduced, and the predicted action distribution becomes more balanced.

These results indicate that LLMs, despite their generative strengths, struggle with action classification, producing biased predictions and exhibiting sensitivity to the context provided in the prompt. Consequently, \textbf{when using LLMs for classification, a well-designed prompt combined with relevant information is crucial for improving accuracy.} \\

\begin{figure}[t]
    \centering
    \vspace{-2mm}
    \includegraphics[width=0.45\textwidth]{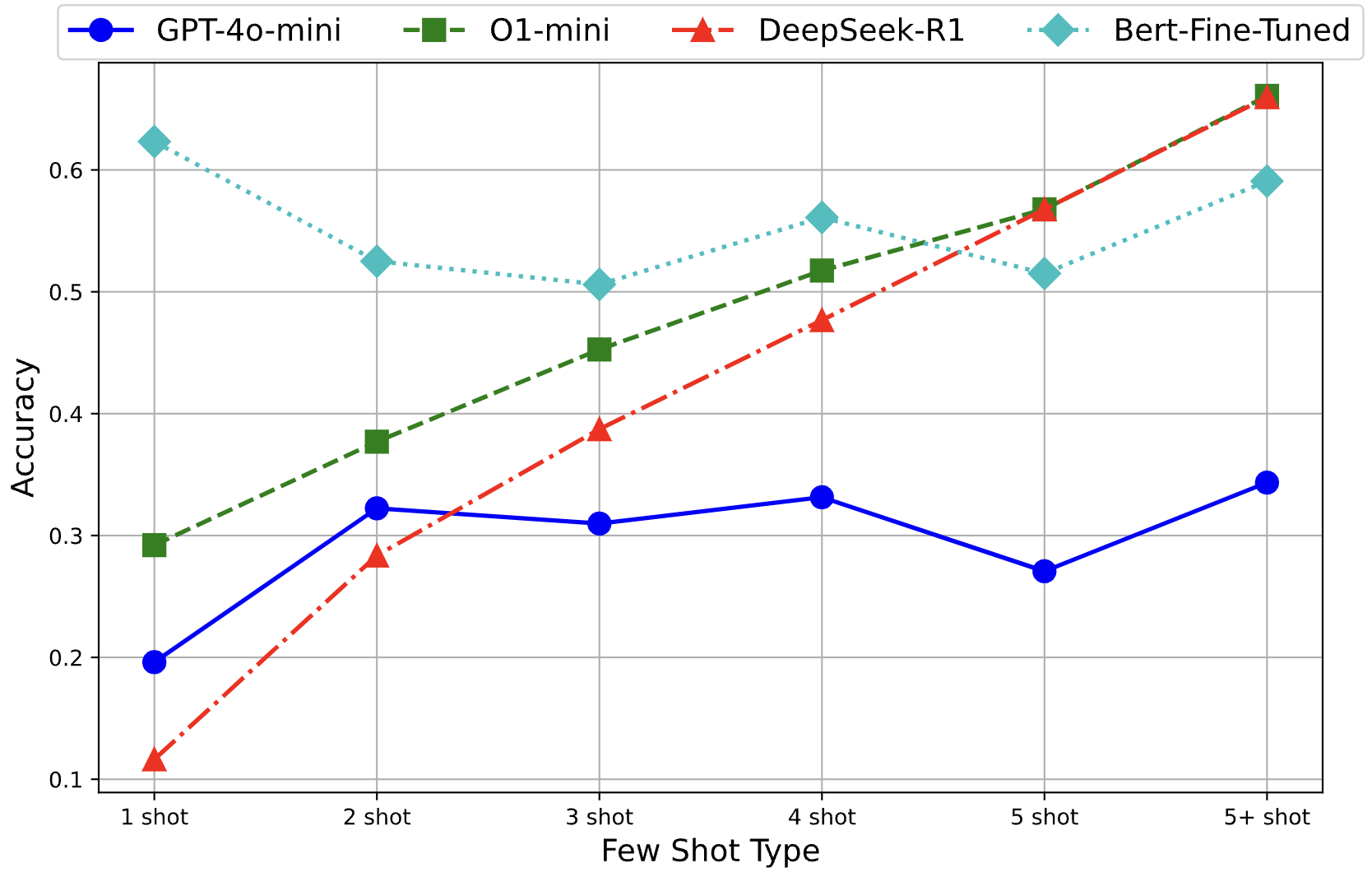}
    \vspace{-2mm}
    \caption{Model accuracy for action prediction across different few-shot settings.}
    \vspace{-3mm}
    \label{fig:accuracy_plot}
\end{figure}

\vspace{-4mm}
\subsection{Few-shot
prompting reduces  accuracy but improves with more examples}
We observe that the FS\_hist configuration achieves lower action prediction accuracy compared to ZS\_info. To investigate the reasons behind this performance degradation, we compare the number of examples in the prompt and prediction accuracy. Since the number of historical examples per user in the dataset is diverse (ranging from 1 to 31), we categorize users into six groups based on the number of historical examples available. 
Figure~\ref{fig:accuracy_plot} presents the action prediction performance of {\tt GPT-4o-mini}, {\tt O1-mini}, {\tt DeepSeek-R1}, and Fine-tuned BERT across different shot configurations, ranging from 1-shot to 5+ shots. We observe that for {\tt O1-mini} and {\tt DeepSeek-R1}, the prediction accuracy is below 30\% at 1-shot but steadily increases as more examples are provided. In contrast, {\tt GPT-4o-mini} shows the lowest accuracy at 1-shot, reaching 19.61\%, which is even lower than its zero-shot performance. In comparison, the BERT model maintains consistently higher accuracy across all shot configurations.
This suggests that using only \textbf{a few examples in prompts can negatively impact LLMs' action prediction accuracy}, likely due to overfitting to a limited context. \\

\vspace{-4mm}
\subsection{Few-shot LLMs achieve stronger semantic alignment with ground-truth posts}
Despite performing worse in action prediction, the few-shot (FS\_hist) LLM setting achieves higher scores in the text generation task compared to zero-shot settings (ZS\_base and ZS\_info). From Table~\ref{tab:action_text_performance}, we observe that FS\_hist consistently outperforms the other configurations in sentiment, emotion, stance, and gender alignment. Specifically, {\tt DeepSeek-R1} under FS\_hist achieves the highest scores for sentiment (68.85\%), emotion (62.88\%), stance (70.05\%), and gender (70.20\%), demonstrating that few-shot prompting enhances semantic alignment between generated responses and ground-truth posts.

It is a known issue that LLMs tend to exhibit bias in age prediction~\cite{kotek2023gender}, which could affect the reliability in this category. 
Regarding stylistic consistency, as measured by style attributes and LIWC scores, we do not observe significant differences across ZS\_base, ZS\_info, and FS\_hist. 

\vspace{-1mm}
\section{Discussions and Conclusions}
In this work, we evaluate LLMs' ability to simulate social media engagement by following an action-guided response generation framework, where a model first predicts a user's engagement type (retweet, quote, or rewrite) before generating a response. 
We benchmark {\tt GPT-4o-mini}, {\tt O1-mini}, and {\tt DeepSeek-R1} on this task. 
Our findings reveal that zero-shot LLMs tend to be biased toward predicting ``quote'' in action prediction, while few-shot LLMs initially reduce prediction accuracy with limited examples. 
This highlights the importance of a well-structured prompt when using LLMs for classification task. Moreover, in response generation, few-shot LLMs produce the most semantically aligned responses with ground truth posts. These insights contribute to a deeper understanding of LLMs' performance on social media simulations, highlighting both their strengths and limitations in modeling realistic user interactions.

\section*{Limitations}
While our findings provide valuable insights into improving AI models that simulate social media interactions, our approach has certain limitations. Action prediction accuracy remains a challenge, particularly in few-shot settings. Additionally, our evaluation is limited to a single social media context, which may affect generalization. Future work should explore combining LLMs with structured classifiers for action prediction, leveraging retrieval-augmented generation (RAG) for better historical context integration, and conducting cross-platform testing to improve adaptability.

\section*{Ethical Considerations}
A potential risk of our study is that LLM-based social media simulations may inadvertently reinforce biases present in the models or training data, leading to skewed representations of user engagement behaviors. Additionally, if misused, such simulations could be leveraged to manipulate online discourse or generate inauthentic engagement patterns, raising ethical concerns about their deployment in real-world applications. To mitigate these risks, we call for increased transparency in LLM-based social media simulations, the development of bias mitigation strategies, and the establishment of ethical guidelines to ensure responsible deployment in research and real-world applications.

\bibliography{reference}

\begin{thebibliography}{21}
\providecommand{\natexlab}[1]{#1}

\bibitem[{Cai et~al.(2024)Cai, Li, Zhang, Li, Wang, and Tei}]{cai2024language}
Jinyu Cai, Jialong Li, Mingyue Zhang, Munan Li, Chen-Shu Wang, and Kenji Tei. 2024.
\newblock Language evolution for evading social media regulation via llm-based multi-agent simulation.
\newblock \emph{arXiv preprint arXiv:2405.02858}.

\bibitem[{Dmonte et~al.(2024)Dmonte, Zampieri, Lybarger, and Albanese}]{dmonte2024classifying}
Alphaeus Dmonte, Marcos Zampieri, Kevin Lybarger, and Massimiliano Albanese. 2024.
\newblock Classifying human-generated and ai-generated election claims in social media.
\newblock \emph{arXiv preprint arXiv:2404.16116}.

\bibitem[{Gao et~al.(2023)Gao, Lan, Lu, Mao, Piao, Wang, Jin, and Li}]{gao2023s}
Chen Gao, Xiaochong Lan, Zhihong Lu, Jinzhu Mao, Jinghua Piao, Huandong Wang, Depeng Jin, and Yong Li. 2023.
\newblock S3: Social-network simulation system with large language model-empowered agents.
\newblock \emph{arXiv preprint arXiv:2307.14984}.

\bibitem[{Jiang and Ferrara(2023)}]{jiang2023social}
Julie Jiang and Emilio Ferrara. 2023.
\newblock Social-llm: Modeling user behavior at scale using language models and social network data.
\newblock \emph{arXiv preprint arXiv:2401.00893}.

\bibitem[{Kotek et~al.(2023)Kotek, Dockum, and Sun}]{kotek2023gender}
Hadas Kotek, Rikker Dockum, and David Sun. 2023.
\newblock Gender bias and stereotypes in large language models.
\newblock In \emph{Proceedings of the ACM collective intelligence conference}, pages 12--24.

\bibitem[{Lim and Schm{\"a}lzle(2023)}]{lim2023artificial}
Sue Lim and Ralf Schm{\"a}lzle. 2023.
\newblock Artificial intelligence for health message generation: an empirical study using a large language model (llm) and prompt engineering.
\newblock \emph{Frontiers in Communication}, 8:1129082.

\bibitem[{Lyu et~al.(2024{\natexlab{a}})Lyu, Huang, Zhang, Yu, Mou, Pan, Yang, Wei, and Luo}]{lyu2024gpt}
Hanjia Lyu, Jinfa Huang, Daoan Zhang, Yongsheng Yu, Xinyi Mou, Jinsheng Pan, Zhengyuan Yang, Zhongyu Wei, and Jiebo Luo. 2024{\natexlab{a}}.
\newblock \href {https://doi.org/10.1145/3709005} {Gpt-4v(ision) as a social media analysis engine}.
\newblock \emph{ACM Trans. Intell. Syst. Technol.}
\newblock Just Accepted.

\bibitem[{Lyu et~al.(2024{\natexlab{b}})Lyu, Pan, Wang, and Luo}]{lyu2024computational}
Hanjia Lyu, Jinsheng Pan, Zichen Wang, and Jiebo Luo. 2024{\natexlab{b}}.
\newblock Computational assessment of hyperpartisanship in news titles.
\newblock In \emph{Proceedings of the International AAAI Conference on Web and Social Media}, volume~18, pages 999--1012.

\bibitem[{Lyu et~al.(2022)Lyu, Wang, Wu, Duong, Zhang, Dye, and Luo}]{lyu2022social}
Hanjia Lyu, Junda Wang, Wei Wu, Viet Duong, Xiyang Zhang, Timothy~D Dye, and Jiebo Luo. 2022.
\newblock Social media study of public opinions on potential covid-19 vaccines: informing dissent, disparities, and dissemination.
\newblock \emph{Intelligent medicine}, 2(01):1--12.

\bibitem[{Miller et~al.(2016)Miller, Sinanan, Wang, McDonald, Haynes, Costa, Spyer, Venkatraman, and Nicolescu}]{miller2016world}
Daniel Miller, Jolynna Sinanan, Xinyuan Wang, Tom McDonald, Nell Haynes, Elisabetta Costa, Juliano Spyer, Shriram Venkatraman, and Razvan Nicolescu. 2016.
\newblock \emph{How the world changed social media}.
\newblock UCL press.

\bibitem[{Mou et~al.(2024)Mou, Wei, and Huang}]{mou2024unveiling}
Xinyi Mou, Zhongyu Wei, and Xuanjing Huang. 2024.
\newblock Unveiling the truth and facilitating change: Towards agent-based large-scale social movement simulation.
\newblock \emph{arXiv preprint arXiv:2402.16333}.

\bibitem[{Pillai et~al.(2025)Pillai, Fokkens, and van Atteveldt}]{pillai2025engagement}
Reshmi~Gopalakrishna Pillai, Antske Fokkens, and Wouter van Atteveldt. 2025.
\newblock Engagement-driven persona prompting for rewriting news tweets.
\newblock In \emph{Proceedings of the 31st International Conference on Computational Linguistics}, pages 8612--8622.

\bibitem[{Qiu et~al.(2024)Qiu, Qiu, Lyu, Xiong, and Luo}]{10825696}
Zhongyi Qiu, Kangyi Qiu, Hanjia Lyu, Wei Xiong, and Jiebo Luo. 2024.
\newblock \href {https://doi.org/10.1109/BigData62323.2024.10825696} {Semantics preserving emoji recommendation with large language models}.
\newblock In \emph{2024 IEEE International Conference on Big Data (BigData)}, pages 7131--7140.

\bibitem[{Rossetti et~al.(2024)Rossetti, Stella, Cazabet, Abramski, Cau, Citraro, Failla, Improta, Morini, and Pansanella}]{rossetti2024social}
Giulio Rossetti, Massimo Stella, R{\'e}my Cazabet, Katherine Abramski, Erica Cau, Salvatore Citraro, Andrea Failla, Riccardo Improta, Virginia Morini, and Valentina Pansanella. 2024.
\newblock Y social: an llm-powered social media digital twin.
\newblock \emph{arXiv preprint arXiv:2408.00818}.

\bibitem[{Roth-Cohen(2022)}]{roth2022viral}
Osnat Roth-Cohen. 2022.
\newblock Viral feminism:\# metoo networked expressions in feminist facebook groups.
\newblock \emph{Feminist Media Studies}, 22(7):1695--1711.

\bibitem[{Salemi et~al.(2025)Salemi, Li, Zhang, Mei, Kong, Chen, Li, Bendersky, and Zamani}]{salemi2025reasoning}
Alireza Salemi, Cheng Li, Mingyang Zhang, Qiaozhu Mei, Weize Kong, Tao Chen, Zhuowan Li, Michael Bendersky, and Hamed Zamani. 2025.
\newblock Reasoning-enhanced self-training for long-form personalized text generation.
\newblock \emph{arXiv preprint arXiv:2501.04167}.

\bibitem[{Toraman et~al.(2022)Toraman, {\c{S}}ahinu{\c{c}}, Yilmaz, and Akkaya}]{toraman2022understanding}
Cagri Toraman, Furkan {\c{S}}ahinu{\c{c}}, Eyup~Halit Yilmaz, and Ibrahim~Batuhan Akkaya. 2022.
\newblock Understanding social engagements: A comparative analysis of user and text features in twitter.
\newblock \emph{Social network analysis and mining}, 12(1):47.

\bibitem[{Van~Dijck and Poell(2013)}]{van2013understanding}
Jos{\'e} Van~Dijck and Thomas Poell. 2013.
\newblock Understanding social media logic.
\newblock \emph{Media and communication}, 1(1):2--14.

\bibitem[{y~Arcas(2022)}]{y2022large}
Blaise~Ag{\"u}era y~Arcas. 2022.
\newblock Do large language models understand us?
\newblock \emph{Daedalus}, 151(2):183--197.

\bibitem[{Yu et~al.(2024{\natexlab{a}})Yu, Li, and Xu}]{yu2024popalm}
Erxin Yu, Jing Li, and Chunpu Xu. 2024{\natexlab{a}}.
\newblock Popalm: Popularity-aligned language models for social media trendy response prediction.
\newblock \emph{arXiv preprint arXiv:2402.18950}.

\bibitem[{Yu et~al.(2024{\natexlab{b}})Yu, Li, and Xu}]{yu2024repalm}
Erxin Yu, Jing Li, and Chunpu Xu. 2024{\natexlab{b}}.
\newblock Repalm: Popular quote tweet generation via auto-response augmentation.
\newblock In \emph{Findings of the Association for Computational Linguistics ACL 2024}, pages 9566--9579.

\end{thebibliography}

\appendix
\section{Appendix}

\subsection{Algorithm Details}
\label{sec:algorithm}
Algorithm~\ref{alg:user_behavior_post_generation} shows the algorithm of our experiments.

\begin{algorithm}[ht]
\caption{User Behavior Simulation for Post Generation}
\label{alg:user_behavior_post_generation}
\begin{algorithmic}[1]
\Require Set of user-trending post pairs $\mathcal{D} = \{(u, p)\}$, classification model $f$, generative model $g$
\Ensure Predicted user actions and generated posts

\For{each $(u, p) \in \mathcal{D}$}
    \State Predict user action: $a \gets f(u, p)$
    \If{$a \in \{\textbf{quote}, \textbf{rewrite}\}$}
        \State Generate new post: $\hat{t} \gets g(u, p)$
    \ElsIf{$a = \textbf{retweet}$}
        \State $\hat{t} \gets \text{None}$ 
    \EndIf
\EndFor

\Return Predicted user actions and generated posts
\end{algorithmic}
\end{algorithm}

\subsection{Attributes and Corresponding Labels}
\label{sec:attributes}
Table~\ref{tab:attributes_labels} shows the attribute and corresponding labels we used to evaluate our models.

\begin{table}[ht]
    \centering
    \caption{Attributes and their corresponding labels used in our evaluation.}
    \begin{tabularx}{\columnwidth}{p{2.5cm}X} 
        \toprule
        \textbf{Attribute} & \textbf{Classification Labels} \\
        \hline
        Sentiment & Positive, Negative, Neutral \\
        Emotion & Sadness, Happiness, Fear, Anger, Surprise, Disgust \\
        Age & Child, Teen, Adult, Senior \\
        Gender & Male, Female \\
        Stance & Favor, Neutral, Against \\
        \bottomrule
    \end{tabularx}
    \label{tab:attributes_labels}
\end{table}

\subsection{Additional Experimental Findings}
\label{sec:additional_exp}

In this section, we present additional insights from our experiments.

\subsubsection{Incorporating User Information Improves Action Prediction Accuracy}

Compared to the Zero-shot Baseline (ZS\_base) approach, adding user information (ZS\_info) significantly enhances action prediction accuracy. This improvement can be attributed to the inclusion of key user-specific features such as retweet rate, quote rate, and rewrite rate, which serve as guidance for the model in selecting the most likely user action.

Furthermore, to better understand the impact of different user information features on action prediction accuracy, we group them into three categories: name (user name), counts (Followers Count, Friends Count, Listed Count, Favourites Count, Statuses Count), and rates (Retweet Rate, Quote Rate, and Rewrite Rate). As shown in Table~\ref{tab:remove_user_info}, we evaluate the model’s performance when specific feature groups are removed from the prompt. 

The results show that removing rates leads to the largest drop in accuracy while removing counts has the smallest effect. Compared to using all features, accuracy decreases by 8.90\% when rates are removed while removing counts results in a 2.28\% increase. This suggests that \textbf{action rates are the most influential factors in action prediction.}

\begin{table}[htbp]
    \centering
    \caption{Impact of removing different user information features on action prediction accuracy.}
    \label{tab:remove_user_info}
    \begin{tabular}{lcc}
        \hline
        \textbf{Features} & \textbf{Accuracy (\%)} & \textbf{Drop (\%)} \\
        \hline
        All Features Used & 42.23 & 0.00 \\
        Name \& Rates & 44.51 & +2.28 \\
        Rates \& Counts & 38.37 & -3.86 \\
        Name \& Counts & 33.33 & -8.90 \\
        \hline
    \end{tabular}
    \vspace{5pt}
    
    \small All user information features include: Name, Followers Count, Friends Count, Listed Count, Favourites Count, Statuses Count, Retweet Rate, Quote Rate, and Rewrite Rate.
\end{table}

\subsubsection{Impact of Action Order on LLMs' Action Prediction}
\begin{figure*}[t]
    \centering
    \includegraphics[width=\textwidth]{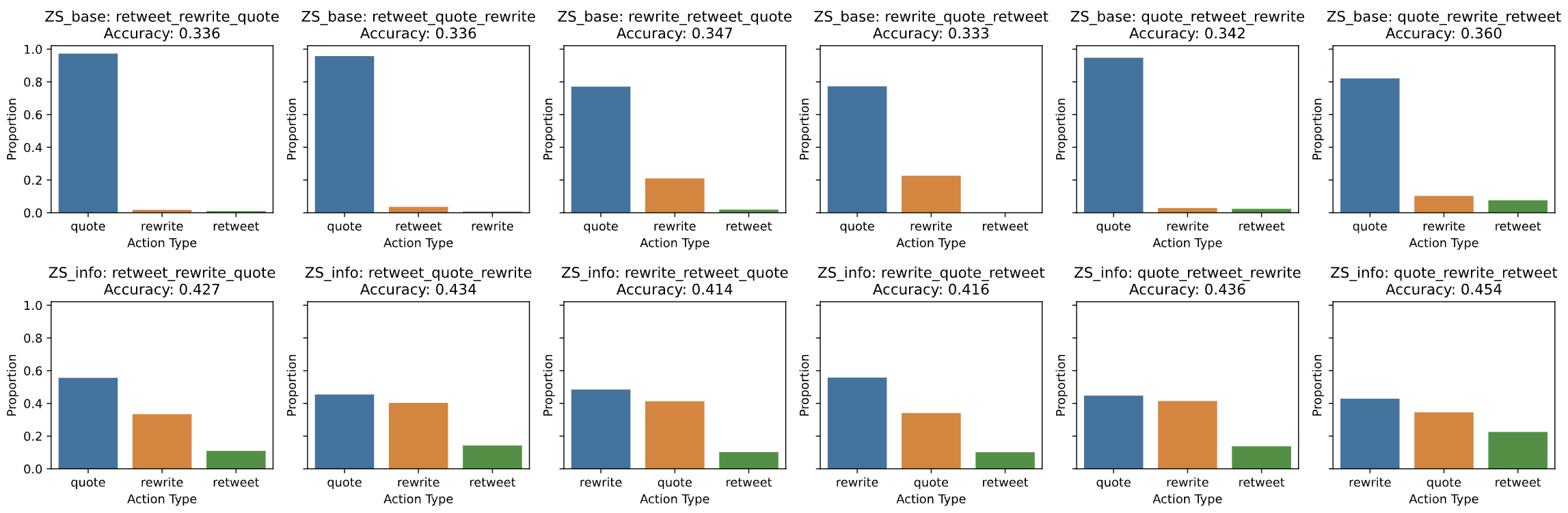}
    \caption{Effect of action order in the prompt on LLM predictions.}
    \label{fig:prompt_order_effect}
\end{figure*}

We analyze how the order of actions (quote, rewrite, retweet) in the prompt affects LLM's action prediction. Figure~\ref{fig:prompt_order_effect} presents the predicted action distributions for different prompt variants, where we alter the order of action labels to test LLMs sensitivity. The result indicates that for ZS\_base and ZS\_info models, reordering the actions has little effect on the predicted distribution.

\end{document}